\documentclass{svproc}
\usepackage{graphicx}
\usepackage{multicol}
\usepackage[bottom]{footmisc}

\usepackage[hidelinks]{hyperref}
\usepackage{placeins}
\usepackage{hhline}
\usepackage{booktabs}                                               
\usepackage{graphicx}  
\usepackage{multirow}                                               
\usepackage{hyphenat}
\usepackage{subcaption}
\usepackage{amsmath}
\usepackage{url}

\usepackage{array}
\usepackage{adjustbox} 
\usepackage{caption}
\begin{document}
\mainmatter              
\title{Improving Medical Waste Classification with Hybrid Capsule Networks}
%
%
\author{Bennet van den Broek \and Javad Pourmostafa Roshan Sharami 
%
%
%
\institute{CSAI Department, Tilburg University, The Netherlands\\
\email{{b.n.n.vdnbroek, J.Pourmostafa}}@tilburguniversity.edu}
}

\maketitle              

\begin{abstract}
The improper disposal and mismanagement of medical waste pose severe environmental and public health risks, contributing to greenhouse gas emissions and the spread of infectious diseases. Efficient and accurate medical waste classification is crucial for mitigating these risks. We explore the integration of capsule networks with a pretrained DenseNet model to improve medical waste classification. To the best of our knowledge, capsule networks have not yet been applied to this task, making this study the first to assess their effectiveness.

A diverse dataset of medical waste images, collected from multiple public sources, is used to evaluate three model configurations: (1) a pretrained DenseNet model as a baseline, (2) a pretrained DenseNet with frozen layers combined with a capsule network, and (3) a pretrained DenseNet with unfrozen layers combined with a capsule network. Experimental results demonstrate that incorporating capsule networks improves classification performance, with F1 scores increasing from 0.89 (baseline) to 0.92 (hybrid model with unfrozen layers). This highlights the potential of capsule networks to address the spatial limitations of traditional convolutional models and improve classification robustness.

While the capsule-enhanced model demonstrated improved classification performance, direct comparisons with prior studies were challenging due to differences in dataset size and diversity. Previous studies relied on smaller, domain-specific datasets, which inherently yielded higher accuracy. In contrast, our study employs a significantly larger and more diverse dataset, leading to better generalization but introducing additional classification challenges. This highlights the trade-off between dataset complexity and model performance. 

Future research should explore fine-tuning model architectures, integrating additional data augmentation techniques, and evaluating the model on standardized benchmark datasets. These efforts will be essential to further optimize the application of capsule networks in medical waste classification and enhance their real-world applicability in automated waste management systems.

\keywords{Medical waste classification, Capsule networks, DenseNet, Transfer learning, Deep learning, Waste management, Environmental sustainability}
\end{abstract}

\section{Introduction} \label{sec:introduction}

In this section, we explore the background of our work, highlighting why it matters both socially and scientifically. We share the main goal of our study, as well as the research questions and hypotheses that guide us. You will also find a brief overview of the key findings we uncovered along the way.

\subsection{Context}

The growing aging population worldwide has led to an increased demand for healthcare services. Consequently, this surge has resulted in a significant rise in unsafe disposal practices and mismanagement of medical waste in healthcare settings \cite{su9020220}. Proper classification of medical waste is crucial for optimizing recycling processes and ensuring safe disposal, both of which are essential for environmental sustainability and public health \cite{Cesaro2017}. Thus, there is a pressing need for accurate methods to classify medical waste, facilitating improved sorting, recycling, and disposal, ultimately reducing environmental impact and minimizing health risks.

Currently, the international organization Health Care Without Harm (HCWH) reports that the healthcare sector contributes 4.4\% to the global climate footprint. Their findings indicate that if the healthcare sector were a country, it would rank as the fifth-largest emitter of greenhouse gases globally, underscoring its environmental impact \cite{karlinerslotterback2024}. Beyond environmental concerns, improperly managed medical waste poses a severe health risk. Each year, over 5.2 million people, including 4 million children, lose their lives due to diseases associated with inadequate medical waste management \cite{dailystar2020}. These alarming statistics highlight the urgency of addressing the growing medical waste crisis.

Advanced technologies, particularly deep learning models, offer promising solutions for medical waste classification. These technologies enable more efficient sorting, recycling, and safe disposal. By leveraging deep learning, healthcare facilities can not only reduce their environmental footprint but also mitigate health risks associated with improper medical waste disposal. One widely studied deep learning model for classification tasks is the Convolutional Neural Network (CNN). However, a significant limitation of CNNs is their difficulty in recognizing objects in images captured from different angles or against varying backgrounds \cite{9591759}. This issue arises primarily due to the pooling layer, which discards spatial relationships between features \cite{Zhang2020EnhancedCN}.

While CNNs have been extensively applied to image classification tasks, limited research has explored the use of capsule networks in medical waste classification. Capsule networks, a relatively new type of neural network, introduce an intermediate structure between neurons and layers (referred to as capsules). Unlike CNNs, capsule networks preserve spatial relationships between features across multiple layers, making them more effective in recognizing transformed objects \cite{rs11141694}.

To date, capsule networks have primarily been applied to general waste classification tasks, where they have demonstrated superior performance compared to CNNs \cite{8728405,huang2023waste}. However, no studies have specifically investigated their application to medical waste classification. Medical waste images present unique challenges not commonly encountered in standard image datasets. These images frequently contain complex objects with irregular shapes and varying sizes, often undergoing transformations. For example, syringes appear in multiple forms and sizes, making accurate classification difficult for traditional CNNs. Capsule networks, however, may offer a more effective solution for handling these complexities.

This study aims to contribute to the existing literature by evaluating the effectiveness of capsule networks in medical waste classification, particularly when combined with a pretrained convolutional neural network. The hybrid model leverages the strengths of both architectures to overcome the limitations of CNNs. The findings of this research are expected to enhance our understanding of capsule networks in real-world scenarios, improving classification performance in medical waste management and encouraging further research into their application for complex imaging tasks. Beyond its scientific contributions, this research may also lead to the development of more reliable and efficient medical waste management systems, ultimately benefiting public health and environmental sustainability.

To achieve this, the study employs a hybrid model integrating a pretrained CNN with a capsule network. Pretrained models, trained on large-scale datasets, provide robust feature extraction capabilities. This enables the capsule network to build upon these features and focus on challenges such as object recognition under varying angles, transformations, and backgrounds, which are typical of medical waste images.

A study by \cite{taher2024medcapsnet} employed a combination of a pretrained DenseNet model and a capsule network for medical heel disease detection. Inspired by this approach, this research adopts a similar hybrid structure. DenseNet was selected as the baseline model due to its optimized information flow between layers, where each layer is directly connected to every other layer. This architecture enhances gradient flow, mitigates the vanishing gradient problem, and improves parameter efficiency \cite{taher2024medcapsnet}.

\subsection{Research Questions}

To address these gaps, the main research question explored in this study is:

\textit{"How can capsule networks improve the classification of medical waste compared to a pretrained DenseNet model as a baseline?"}

To further investigate this, the following subquestions are formulated:

\begin{itemize}
    \item \textbf{Subquestion 1.1:} \textit{“How does the performance of a pretrained DenseNet with fine-tuned layers compare to that of a pretrained DenseNet combined with a capsule network, in terms of accuracy, precision, recall, and F1-score?”}
    \item \textbf{Subquestion 1.2:} \textit{“How does the performance of a pretrained DenseNet used solely as a feature extractor (with frozen layers) compare to a pretrained DenseNet with unfrozen layers, both combined with a capsule network, in terms of accuracy, precision, recall, and F1-score?”}
\end{itemize}

It is hypothesized that combining a capsule network with a pretrained model will enhance classification performance, as capsule networks preserve spatial relationships between features. This ability makes them particularly well-suited for medical waste images, which contain complex variations in appearance and positioning. Furthermore, it is expected that the model with unfrozen layers will outperform the model with frozen layers, as fine-tuning allows the pretrained model to adapt to the dataset’s specific characteristics, thereby improving its effectiveness.

\subsection{Main Findings}

This study contributes to the existing literature in several ways:

\begin{enumerate}
    \item It represents the first documented use of capsule networks for medical waste classification.
    \item Given the relatively unexplored nature of capsule networks, this study provides valuable insights into their potential for handling complex classification tasks.
    \item The results indicate that capsule networks enhance performance when integrated with a pretrained DenseNet model for medical waste classification. Compared to previous studies, the overall performance remains lower, likely due to differences in dataset size and composition. Prior work used smaller, domain-specific datasets, leading to higher accuracy, whereas this study employs a larger, more diverse dataset that improves generalization but introduces additional classification challenges. 
\end{enumerate}

This research assesses whether hybrid capsule networks can overcome CNN limitations, particularly in recognizing objects under varying angles, backgrounds, and transformations. Ultimately, the goal is to demonstrate how this hybrid model can improve classification performance, contributing to more efficient sorting, recycling, and safe disposal of medical waste in healthcare settings.

\section{Related Work}

The need for effective medical waste classification has grown in recent years due to increasing concerns over environmental sustainability and public health risks. Several studies have explored deep learning techniques to address this challenge, providing valuable insights into their applications.

To offer a structured overview of previous research, the literature is categorized into three main areas. The first category examines studies that apply deep learning models, particularly CNNs, to medical waste classification. The second category explores research on capsule networks for general waste classification, as their application in medical waste classification remains unexplored. The third category integrates these insights, demonstrating how this research builds upon and contributes to existing knowledge by applying capsule networks to medical waste classification.

\subsection{Using Deep Learning Models for Medical Waste Classification}

Several studies have investigated the use of deep learning models for medical waste classification. For instance, \cite{10119431} explored multiple deep learning models, including EfficientNetB7, VGG19, InceptionV3, and ResNet50, combined with transfer learning. Their model classified medical waste into three categories: general, hazardous, and infectious. Three of the four models achieved accuracy rates exceeding 95\%, demonstrating the strong performance of deep learning techniques in medical waste classification. \cite{9855374} classified medical waste into two categories: sharp infectious waste and non-sharp infectious waste. Using five different deep learning models—VGG16, VGG19, ResNet50, InceptionV3, and MobileNetV2—combined with data augmentation, they achieved an impressive 99.40\% accuracy with ResNet50. While these studies highlight the potential of deep learning in medical waste classification, they are limited to simplified categorizations––grouping waste into broad classes without distinguishing between specific item types.

A more specialized approach was taken by \cite{zhou2022deep}, who applied a deep learning model, ResNeXt, enhanced with transfer learning, to classify eight types of medical waste, including gauze, gloves, infusion bags, syringes, and needles. Using a dataset of 3,480 images and applying data augmentation to mitigate overfitting, they achieved an accuracy of 97.2\%. \cite{refId0} classified seven different types of medical waste, including vials, masks, syringes, gloves, cotton, bandages, and IV tubes. Their study, conducted in developing countries, particularly in Thailand, employed the EfficientNetB7 model with transfer learning, achieving a remarkable accuracy of 99\%. However, concerns arise regarding the robustness of these results due to the small test set, which contained only 337 images. Although these studies report high accuracy, their reliance on relatively small datasets raises concerns about their generalizability and the potential risk of overfitting.

\subsection{Using Capsule Networks for \textit{General} Waste Classification}

Capsule networks, a relatively new deep learning approach, have received limited research attention across different classification fields. However, a few studies have examined their use in general waste classification. One of the earliest studies in this domain, conducted by \cite{8728405}, compared the performance of capsule networks with traditional CNNs for classifying plastic and non-plastic waste. Their datasets, comprising over 10,000 and 5,000 images, ensured robust results. The capsule network achieved accuracies of 96.3\% and 95.7\% on the two datasets, outperforming CNNs, which achieved slightly lower accuracies of 95.8\% and 93.6\%. This study established a key benchmark, demonstrating the ability of capsule networks to preserve spatial relationships and texture details—an area where CNNs often struggle.

Expanding on this, \cite{janeera2021visual} classified five categories of general waste, including paper, plastic, metal, non-metal, and uncategorizable waste. They implemented a hybrid approach, combining a CNN for feature extraction with a capsule network for classification. This method leveraged the strengths of both models, achieving an impressive accuracy of 98.2\%. Despite the relatively small dataset used, their findings underscore the potential of capsule networks for multi-category waste classification. A more recent study by \cite{huang2023waste} investigated the application of capsule networks for classifying four types of waste: glass, metal, paper, and plastic. Using a dataset of approximately 2,000 images, they adopted a hybrid model that combined Residual Networks (ResidualNet) with capsule networks. This approach outperformed several deep learning models, including AlexNet, VGG16, and ResNet18. The capsule network model achieved a classification accuracy of 91\%, compared to 83\%, 82\%, and 86\% for AlexNet, VGG16, and ResNet18, respectively.

\subsection{Using Capsule Networks for \textit{Medical} Waste Classification}

The existing literature demonstrates the effectiveness of deep learning models in medical waste classification, with many studies achieving high accuracy. However, most of these studies either focus on broad categorizations or rely on small datasets, raising concerns about the robustness and generalizability of their models. Although capsule networks have shown promising results in general waste classification, their application to medical waste classification remains unexplored. These networks excel in preserving spatial and texture information, as evidenced by studies such as \cite{8728405}. Traditional CNNs struggle in capturing these fine-grained features and spatial hierarchies, making capsule networks a promising alternative. Given that medical waste classification involves distinguishing between objects such as syringes, gloves, masks, and medical vials—items that vary significantly in shape, texture, and orientation—capsule networks could offer significant advantages in improving classification performance.

By addressing this research gap, we aim to enhance the accuracy and reliability of medical waste classification. The findings have the potential to contribute to more efficient medical waste management systems, which are crucial for reducing environmental impact and minimizing health risks. As demonstrated in previous research, such as~\cite{zhou2022deep}, combining a pretrained model with a capsule network often enhances classification performance. This study follows a similar approach to evaluate the benefits of integrating capsule networks into medical waste classification.

\section{Data Collection and Processing}
This section outlines the datasets utilized in the study, along with preprocessing steps and data augmentation techniques to address class imbalance.

\subsection{Dataset Creation}
\label{sec:data}

\begin{figure}[!ht] \centering \includegraphics[width=1\linewidth]{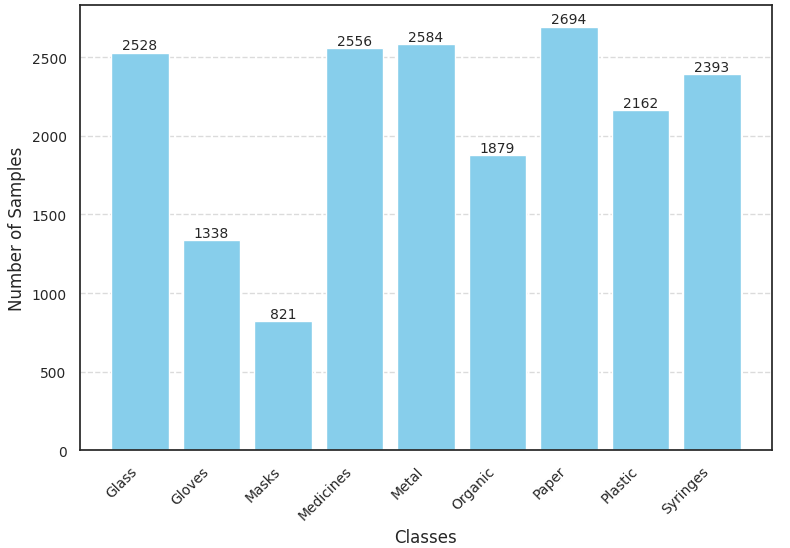} \caption{Class distribution before data augmentation} \label{fig:before} \end{figure}

To enhance the generalizability of the model, a large and diverse set of medical waste images is essential. However, due to limited data availability, multiple datasets are combined to increase the number of images. The primary dataset, The Trashbox~\cite{kumsetty_2022}, includes approximately 500 images each of medical waste items such as syringes, surgical gloves, surgical masks, and medicines. Also, it contains around 2,500 images each of plastic, paper, metal, and glass. To further expand the dataset, we incorporate a dataset by \cite{abdullah_2024}, which includes approximately 1,000 images per category of glass, paper, plastic, and organic waste. Additionally, a dataset from \textit{Images.cv}~\cite{syringe_dataset_2024} contributes 2,000 syringe images, while another Kaggle dataset by~\cite{radli_2024} provides 2,000 images of medicines. Integrating these datasets enhances diversity and ensures a more comprehensive representation of medical waste categories. However, class imbalance occurs because of the unequal distribution of images, which we addressed through data augmentation. Figure~\ref{fig:before} illustrates the class distribution before data augmentation.

\noindent After dataset creation, preprocessing steps are applied to ensure consistency and improve model performance. All images are resized to 224×224 pixels to standardize input dimensions. To maintain aspect ratios and prevent distortion, black padding is added where necessary. The dataset is then split into training (70\%), validation (15\%), and test (15\%) sets while preserving class distribution.

\subsection{Data Augmentation}
\begin{figure}[!ht]
    \centering
    \includegraphics[width=1\linewidth]{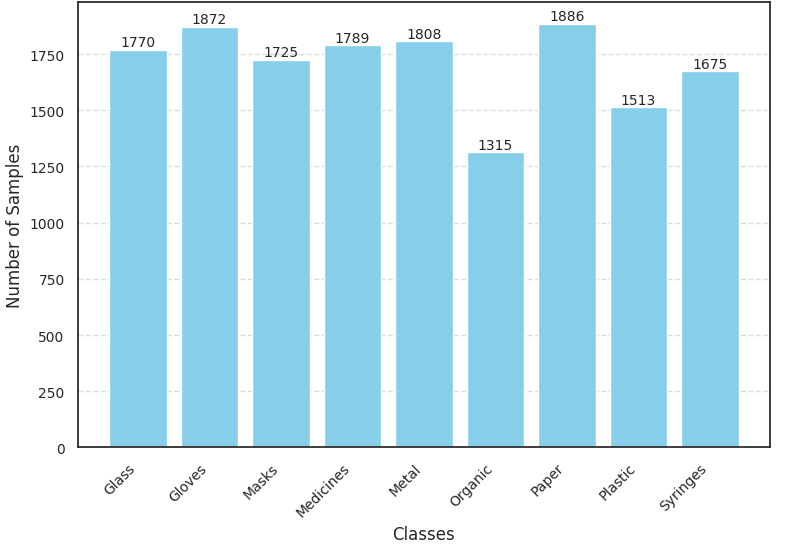}
    \caption{Class distribution after data augmentation}
    \label{fig:after}
\end{figure}

 To mitigate class imbalance in the training data, augmentation techniques such as rotation and scaling are applied to underrepresented classes, particularly gloves and masks. Specifically, one additional image is generated for each glove image, while two additional images are created for each mask image. The resulting class distribution after augmentation is shown in Figure~\ref{fig:after}.

\subsubsection{Data Normalization and Formatting}

After data augmentation, all images are normalized by scaling pixel values to the range [0, 1]. Class labels are converted into a one-hot encoded format for accurate multi-class classification. Additionally, since the capsule network is implemented with PyTorch, the dataset is converted into PyTorch tensors. To ensure efficient data loading during training, validation, and testing, PyTorch DataLoader objects are created for each dataset. These objects manage batching, shuffling, and organized data processing, optimizing the dataset for model input. The capsule network model in use is implemented with the PyTorch library, so the dataset must be converted to PyTorch tensors. To enable efficient data loading during model training, validation, and testing, PyTorch DataLoader objects are created for each dataset. These datasets package the input data and labels as tensors, ensuring that the data is well-organized, shuffled, and ready for batch processing. Afterward, the data is ready to serve as input for the model.

The dataset used in our experiments is publicly accessible at: \url{https://github.com/BennetvdBroek/Thesis-Capsule-Network-code}.  

\section{Models}

\subsection{Baseline––DenseNet121}

DenseNet is a convolutional neural network comprising multiple dense blocks, where each layer is connected to all previous layers. This design encourages feature reuse, reduces redundancy, and enhances gradient flow, making the model more efficient and less prone to overfitting. Additionally, each layer receives direct supervision from the loss function, improving optimization efficiency~\cite{zhu2017densenet}.

For this study, we employ \textit{DenseNet121}, a 121-layer variant of DenseNet. A subset of its pretrained layers is unfrozen, treating the number of trainable layers as a hyperparameter. To adapt the extracted features for classification, we introduce a flattening layer (from PyTorch) followed by fully connected layers, with dropout applied to mitigate overfitting. \textit{This configuration serves as our baseline model.}

\subsection{Capsule Networks}
Conventional CNNs struggle with capturing spatial hierarchies, leading to limitations when handling rotated, tilted, or differently positioned objects. Capsule networks, introduced by~\cite{sabour2017dynamic}, address this limitation by encoding object properties through vectorized neuron groups called capsules. Unlike CNNs, which discard spatial relationships in pooling layers, capsule networks preserve orientation, scale, and positional details~\cite{PAWAN2022102}. 

We utilize capsule networks as the primary model architecture for classification, leveraging their ability to retain spatial information and improve robustness to object transformations.

\subsection{Hybrid Model: DenseNet-Capsule Network}
To assess the effectiveness of capsule networks for medical waste classification, we propose a hybrid approach that integrates \textit{DenseNet121} with a capsule-based classification mechanism. Following the methodology of~\cite{taher2024medcapsnet}, we evaluate two hybrid configurations:

\begin{enumerate}
    \item \textbf{Frozen DenseNet121:} The DenseNet121 model serves purely as a feature extractor, with all layers frozen. Its output is passed to the capsule network for classification.
    \item \textbf{Fine-tuned DenseNet121:} Some layers of DenseNet121 are unfrozen, allowing trainable weights to adapt to the dataset. Dropout is applied to mitigate overfitting.
\end{enumerate}

\subsubsection{Performance Evaluation Metrics}
The performance of the models in this study is evaluated using \textit{accuracy}, \textit{precision}, \textit{recall}, and \textit{F1 score}.

\section{Experiments}
\label{sec:exp}

This study conducts three key experiments to evaluate the effectiveness of different model configurations for medical waste classification. Each experiment builds upon the previous one, progressively incorporating more advanced techniques to assess their impact on classification performance.

\subsection{Experiment 1: Baseline Model – Pretrained DenseNet121}

In the first experiment, a pretrained DenseNet121 model is fine-tuned for medical waste classification. This serves as the baseline for comparison with more advanced models. To prevent overfitting, only the final layers of the DenseNet121 model are unfrozen, allowing them to be trainable while leveraging the pretrained convolutional base.

The model architecture consists of the DenseNet121 feature extractor, followed by a flattening layer to reshape the extracted features. To further reduce overfitting, L2 regularization is applied to the trainable weights, and dropout is introduced. The final layer is a fully connected dense layer with 9 units and a SoftMax activation function, producing a probability distribution across the 9 classes. The best-performing weights from this baseline model are saved for test set evaluation.

\subsection{Experiment 2: Pretrained DenseNet121 with Frozen Layers}

In the second experiment, the pretrained DenseNet121 model is used with all layers frozen, meaning that the weights remain unchanged during training and are solely used for feature extraction. The extracted features serve as the input to a capsule network, forming a hybrid model that integrates DenseNet121 and capsule networks.

\subsection{Experiment 3: Pretrained DenseNet121 with Unfrozen Layers}

In the final experiment, the same pretrained DenseNet121 model is used, but with all layers unfrozen to the same extent as in the baseline model. This allows the model to fine-tune its weights during training, adapting to the specific characteristics of the medical waste dataset. To mitigate overfitting, dropout is incorporated, similar to the baseline model.

This setup ensures that the comparison between the hybrid model and the baseline is fair, allowing for a clear assessment of the impact of integrating capsule networks. The output from the pretrained DenseNet121 model serves as the input to the capsule network component. \\

\noindent Hyperparameter tuning was performed for all three experiments, evaluating multiple configurations to identify the optimal settings. The tested hyperparameters and their corresponding values are summarized in Appendix~\ref{app:b}, with the best-performing configuration selected for final model training. The results of this tuning process are discussed in Section~\ref{sec:results}.\\

\begin{figure}[!ht]
    \centering
    \includegraphics[width=0.9\linewidth]{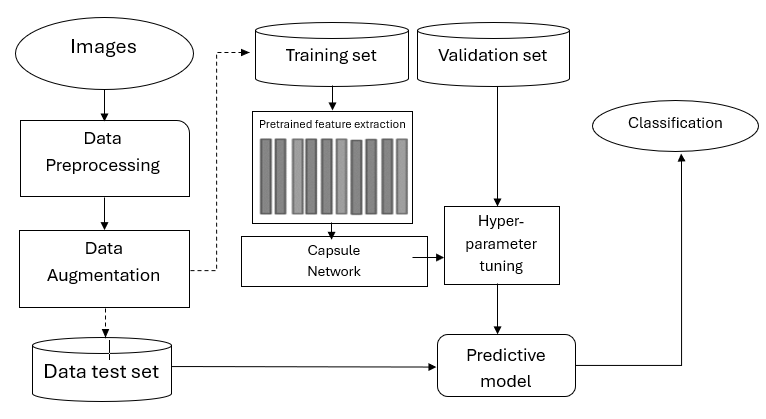}
    \caption{Methodological pipeline detailing feature extraction and classification steps.}
    \label{fig:flowchart}
\end{figure}

Figure~\ref{fig:flowchart} illustrates the methodological pipeline of this study, covering the entire process from input images to classification. It includes key stages such as data preprocessing, augmentation, and hyperparameter tuning.

\subsubsection{Reproducibility and Code Availability}  

The dataset used in this study is publicly available on GitHub, Kaggle, and Images.cv, containing no personal or sensitive data. The full implementation of this study, including all code, is accessible at \url{https://github.com/BennetvdBroek/Thesis-Capsule-Network-code}. Additional details on the software environment, libraries, and computational resources used for reproducibility are provided in Appendix~\ref{app:a}.  

\section{Results}
\label{sec:results}
This section presents the experimental results of medical waste classification using various model configurations. The models were trained on the training set, hyperparameters were adjusted using the validation set, and the final evaluation was conducted on the test set.

The results for each model/experiment are summarized in Table~\ref{tab:merged_results}, allowing for direct comparison across the baseline model, the frozen DenseNet121 with a capsule network, and the unfrozen DenseNet121 with a capsule network.

\subsection{Classification Performance}
Hyperparameter tuning was conducted for all three experiments, testing multiple configurations to determine the optimal settings. Appendix~\ref{app:b} provides a summary of the tested hyperparameters and their values, with the best-performing configuration selected for final model training.

Table~\ref{tab:merged_results} presents the classification results of all models evaluated on the test set. The baseline DenseNet121 model achieved strong performance, with an F1-score of 0.89. Incorporating a capsule network with a frozen DenseNet121 slightly improved the results, increasing the F1-score to 0.896. The best performance was observed when combining the capsule network with an unfrozen DenseNet121 (Experiment 3), achieving an F1-score of \textit{0.918}.

\begin{table}[!ht]
\centering
\renewcommand{\arraystretch}{1.2} 
\setlength{\tabcolsep}{8pt} 

\adjustbox{max width=\textwidth}{
\begin{tabular}{|l|c|r|r|r|}
\hline
\textbf{Model / Experiment} & \textbf{Metric} & \textbf{Precision} & \textbf{Recall} & \textbf{F1-score} \\ \hline
\multirow{2}{*}{Baseline (DenseNet121) / Exp1}       
               & Macro Avg       & 0.889              & 0.894           & 0.891             \\ \cline{2-5}
               & Weighted Avg    & 0.892              & 0.891           & 0.891             \\ \hline
               
\multirow{2}{*}{Frozen DenseNet121 + Capsule / Exp2}  
               & Macro Avg       & 0.894              & 0.897           & 0.896             \\ \cline{2-5}
               & Weighted Avg    & 0.897              & 0.897           & 0.896             \\ \hline
               
\multirow{2}{*}{Unfrozen DenseNet121 + Capsule / Exp3}  
               & Macro Avg       & \textbf{0.917}              &\textbf{ 0.914}           & \textbf{0.918}             \\ \cline{2-5}
               & Weighted Avg    & \textbf{0.916}              & \textbf{0.915}           &\textbf{0.918}             \\ \hline
\end{tabular}
}
\caption{Comparison of classification performance for all models in the test set. ``Exp" indicates the experiment number. Detailed explanations of the experiments and their setups are available in Section~\ref{sec:exp}.}
\label{tab:merged_results}
\end{table}

\subsection{Error Analysis Using Confusion Matrices}

\begin{figure}[!ht]
    \centering
    \includegraphics[width=0.7\linewidth]{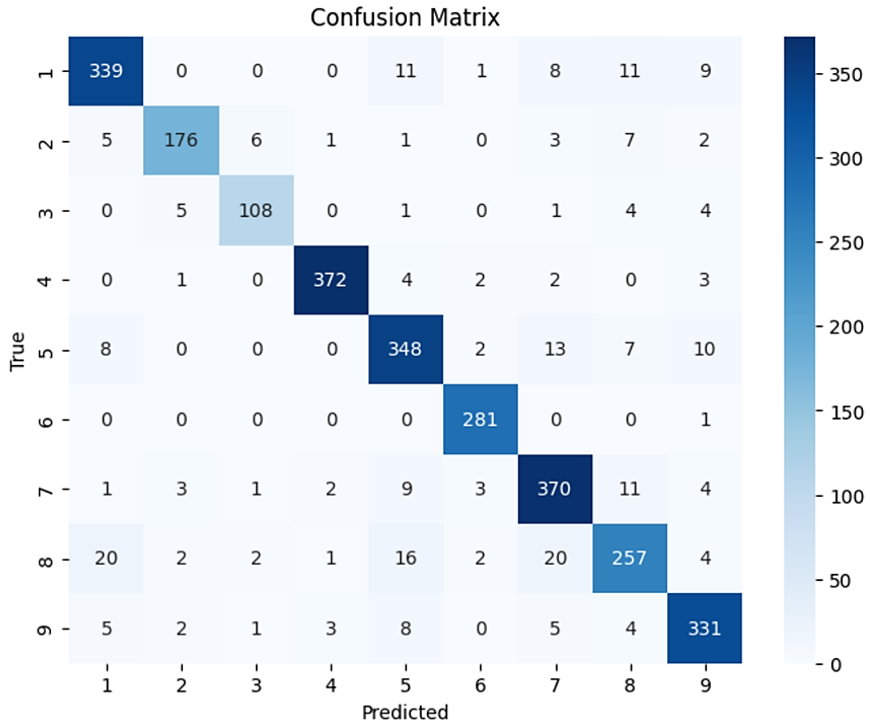}
    \caption{Confusion matrix of the baseline model (DenseNet121).}
    \label{fig:pretrained_cm}
\end{figure}

We used confusion matrices to analyze classification errors and identify patterns of misclassification. Figure~\ref{fig:pretrained_cm} shows that while the baseline model performed well overall, it struggled to differentiate between plastic, glass, and metal, suggesting that the model had difficulty capturing fine-grained features needed to distinguish materials with similar textures.

\begin{figure}[!ht]
    \centering
    \includegraphics[width=0.7\linewidth]{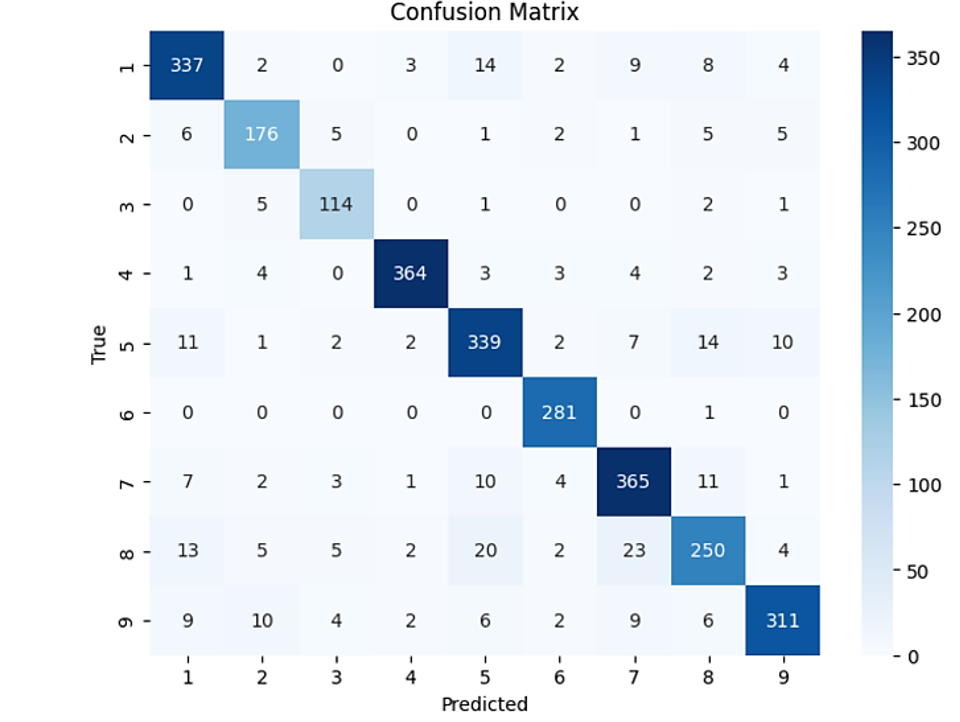}
    \caption{Confusion matrix of the frozen DenseNet121 with the capsule network.}
    \label{fig:frozen_cm}
\end{figure}

The introduction of a capsule network with a frozen DenseNet121 (Figure~\ref{fig:frozen_cm}) led to a slight improvement in distinguishing plastic from other materials. This suggests that the capsule network helped preserve spatial relationships and improved feature representation, making it easier for the model to differentiate between similar-looking objects.

\begin{figure}[!ht]
    \centering
    \includegraphics[width=0.7\linewidth]{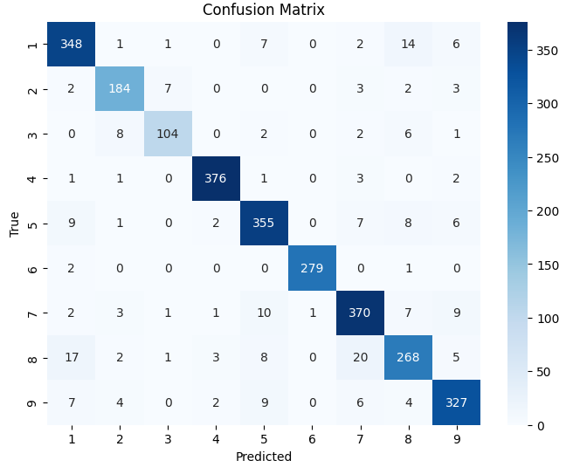}
    \caption{Confusion matrix of the unfrozen DenseNet121 with the capsule network.}
    \label{fig:unfrozen_cm}
\end{figure}

The best performance was achieved with the unfrozen DenseNet121 and capsule network combination (Figure~\ref{fig:unfrozen_cm}), which demonstrated the highest accuracy in separating these materials. The ability to fine-tune the DenseNet121 layers alongside the capsule network allowed the model to extract more relevant features, reducing misclassification errors significantly.

\subsection{Model Comparison and Findings}

The results indicate that integrating a capsule network improved classification performance. The frozen DenseNet121 hybrid model provided a small improvement over the baseline, but the unfrozen version achieved the highest performance, demonstrating the benefits of allowing the model to fine-tune its feature representations.

The confusion matrices show that the unfrozen hybrid model significantly improved differentiation between plastic and other classes, reducing misclassifications observed in previous models. These findings suggest that combining a pretrained model with a capsule network enhances classification accuracy in medical waste categorization.

\noindent For a detailed view of the model’s training behavior, including loss and accuracy trends over epochs, refer to Appendix~\ref{app:training_curves}.

\section{Discussion and Conclusion}

This section presents a comprehensive analysis of the findings, highlighting their significance in relation to existing research. It also discusses study limitations, potential future research directions, and the broader societal implications of our work.

\subsection{Addressing the Research Questions}

The main research question of our study is:  
\textit{"How can capsule networks improve the classification of medical waste compared to a pretrained DenseNet model as a baseline?"}  
To answer this, two subquestions were examined.

The first subquestion investigates the impact of adding a capsule network to a pretrained DenseNet model with the last 30 layers unfrozen. The results indicate that incorporating a capsule network enhances classification performance across all metrics, with the hybrid model achieving an F1-score of approximately 0.92 compared to 0.89 for the baseline model. This confirms that the capsule network is a valuable addition, improving the model’s ability to capture spatial relationships and better classify complex medical waste categories.

However, when compared to prior studies, both the baseline and hybrid models were not tested on the same datasets as other deep learning models used for medical waste classification. For example, studies by \cite{10119431,9855374,zhou2022deep,refId0} reported accuracy rates between 0.95 and 0.99 on their respective datasets, which are not directly comparable to this study. A key reason for this difference could be dataset composition. Unlike prior research that used smaller, domain-specific datasets, this study employed a larger and more diverse dataset, which enhances generalization but may introduce additional classification challenges. These findings highlight the trade-off between dataset diversity and model performance, and future work could explore evaluating the proposed model on these benchmark datasets for a more direct comparison.

The second subquestion explores whether a capsule network performs better when paired with a fully frozen pretrained model or when some layers are trainable. The results indicate that the model with 30 unfrozen layers performed best, suggesting that allowing the pretrained model to fine-tune its feature representations enhances the capsule network’s effectiveness. This aligns with findings from \cite{janeera2021visual} and \cite{huang2023waste}, who also found that trainable pretrained models yielded better classification results.





\subsection{Limitations and Future Research} 

Despite the promising results, several limitations must be considered. Although the dataset used in our work is \textit{larger and more diverse} than those in prior research, it may still not fully represent the vast variety of medical waste encountered in real-world settings. Future studies could explore even more \textit{diverse and larger datasets} to enhance generalizability.

Additionally, only \textit{two data augmentation techniques} were applied in our work. Expanding the range of augmentation techniques could provide further improvements in classification performance, particularly for underrepresented waste categories.

Further \textit{fine-tuning the capsule network architecture} may also lead to performance gains. Exploring \textit{different capsule configurations, additional routing algorithms, or alternative deep learning architectures} could provide further insights into the optimal setup for medical waste classification. Another avenue for improvement is \textit{combining capsule networks with alternative pretrained models}, potentially enhancing feature extraction capabilities.

\subsection{Potential for Real-World Applications}

Future research should also consider the \textit{practical implementation} of capsule networks in \textit{real-time medical waste management systems}. Investigating how these models perform in \textit{live deployment environments} could provide valuable insights into their robustness and scalability.

For instance, \textit{integrating capsule networks into hospital waste management systems} could enhance sorting accuracy, streamline recycling processes, and support sustainable waste disposal practices. Evaluating these models in \textit{real-world scenarios} would be crucial in assessing their feasibility for large-scale deployment.

\section{Conclusion}

We investigated the potential of capsule networks to improve medical waste classification compared to a pretrained DenseNet model. The results suggest that incorporating capsule networks enhances classification performance, with the best-performing hybrid model achieving an F1-score of 0.92. While these results are promising, a direct comparison with state-of-the-art models in the literature is challenging due to differences in datasets and experimental setups. Previous studies \cite{10119431,9855374,zhou2022deep,refId0} reported higher accuracy scores on their respective datasets, which may not generalize in the same way to a more diverse dataset like the one used in this work. This highlights the importance of dataset composition when evaluating model performance and suggests that future work could explore testing the proposed model on benchmark datasets for a more direct comparison.

Nevertheless, our work contributes to the growing field of \textit{medical waste classification} by demonstrating the effectiveness of capsule networks in this domain. It also highlights the importance of allowing pretrained models to fine-tune their feature representations to maximize classification performance.

While further optimizations are required, this research provides a solid foundation for future studies exploring \textit{hybrid deep learning models} for waste classification. Key areas for improvement include \textit{expanding dataset diversity, applying additional augmentation techniques, and optimizing capsule network architectures}.

Beyond its academic contributions, this research holds \textit{practical significance} in \textit{automated waste management and environmental sustainability}. As healthcare facilities worldwide generate increasing amounts of medical waste, integrating capsule networks into real-time waste sorting systems could revolutionize \textit{waste classification, recycling efficiency, and sustainable disposal practices}.

In conclusion, while challenges remain, this study provides a \textit{strong basis for future research} into the application of capsule networks in medical waste classification. Exploring their real-time integration into waste management systems could drive significant advancements in healthcare sustainability and waste reduction.

\section*{Appendix A – Software and Computational Resources}  
\label{app:a}  

All experiments are conducted in Python on Google Colaboratory Pro, utilizing high-performance GPUs and TPUs for efficient model training. The libraries used in the experiments are listed in Table~\ref{tab:libraries}.  

\begin{table}[!ht]
\centering
\begin{tabular}{|l|l|l|}
\hline
\textbf{Library} & \textbf{Version} & \textbf{Source} \\ \hline
NumPy            & 1.26.4           & \cite{harris2020array} \\ \hline
Pandas           & 2.2.2            & \cite{mckinney-proc-scipy-2010} \\ \hline
Matplotlib       & 3.8.0            & \cite{Hunter:2007} \\ \hline
Scikit-learn     & 1.5.2            & \cite{pedregosa2011scikit} \\ \hline
Keras            & 3.5.0            & \cite{chollet2015keras} \\ \hline
TensorFlow       & 2.17.1           & \cite{tensorflow2015-whitepaper} \\ \hline
OpenCV           & 4.10.0           & \cite{itseez2015opencv} \\ \hline
PyTorch          & 2.5.1            & \cite{NEURIPS2019_9015} \\ \hline
\end{tabular}
\caption{Libraries and their versions used in our study.}
\label{tab:libraries}
\end{table}

\section*{Appendix B – Hyperparameter Selection}  
\label{app:b}  

For all our experiments, various hyperparameters were tested, and the highest-performing configurations were selected. The tested hyperparameters and their corresponding values are summarized in Table~\ref{tab:hyperparameters}.  

\begin{table}[!ht]
\centering
\begin{tabular}{|>{\raggedright\arraybackslash}m{5cm}|>{\raggedright\arraybackslash}m{6cm}|}
\hline
\textbf{Hyperparameter} & \textbf{Values} \\ \hline
Optimizer & Adam, RMSprop \\ \hline
Learning rates & 0.01, 0.0001, 0.00001 \\ \hline
Epochs & 100 (early stopping at 10 epochs) \\ \hline
Dropout rates & 0.4, 0.6, 0.8 \\ \hline
L2 regularization weight & 0.001, 0.0001, 0.00001 \\ \hline
Batch size & 50, 100, 150 \\ \hline
Kernel size of primary capsules & 2, 3, 5 \\ \hline
Stride of primary capsules & 1, 2 \\ \hline
Number of frozen layers & 10, 20, 30 \\ \hline
\end{tabular}
\caption{Hyperparameters tested and their respective values.}
\label{tab:hyperparameters}
\end{table}

\section{Appendix C – Training Performance Analysis}
\label{app:training_curves}

This section presents the loss and accuracy curves for all three models during training, providing insights into model convergence and overfitting tendencies.

\begin{figure}[!ht]
    \centering
    \includegraphics[width=1\linewidth]{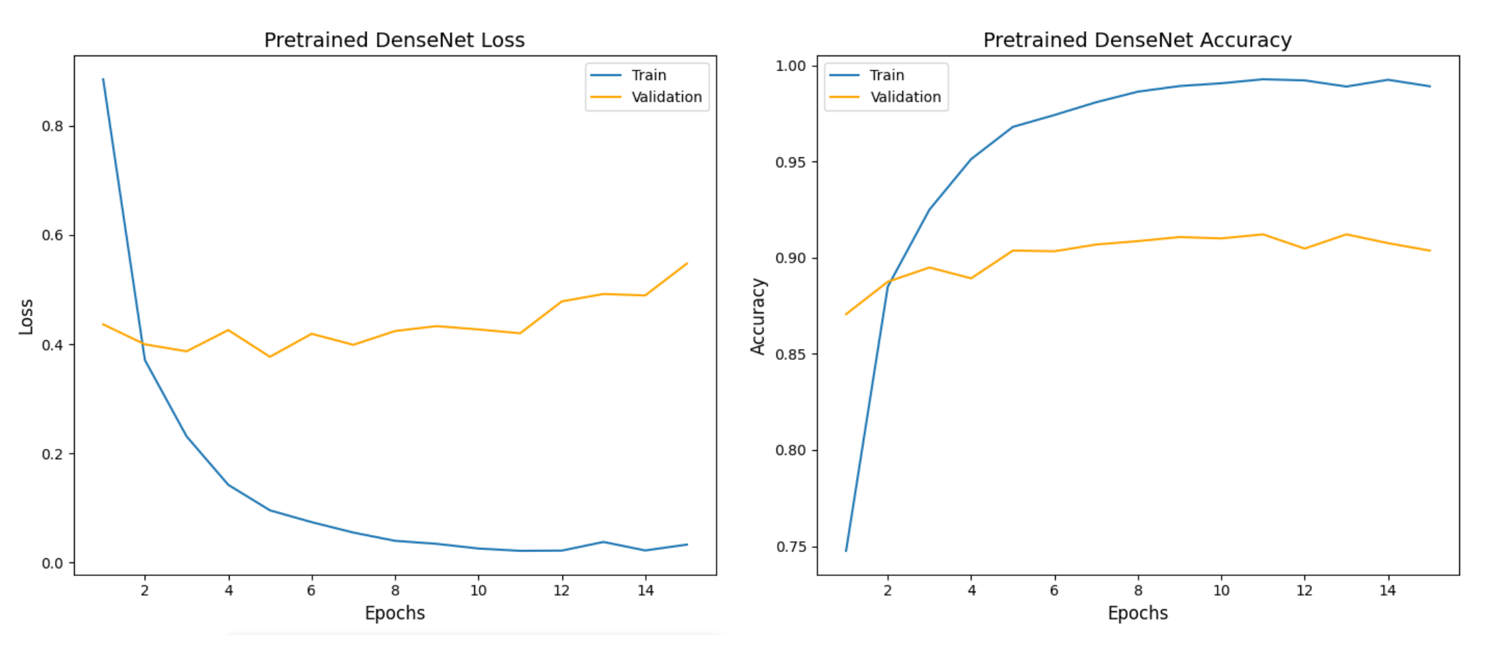}
    \caption{Loss and accuracy curves of the baseline model (DenseNet121).}
    \label{fig:pretrained_loss_acc}
\end{figure}

\begin{figure}[!ht]
    \centering
    \includegraphics[width=1\linewidth]{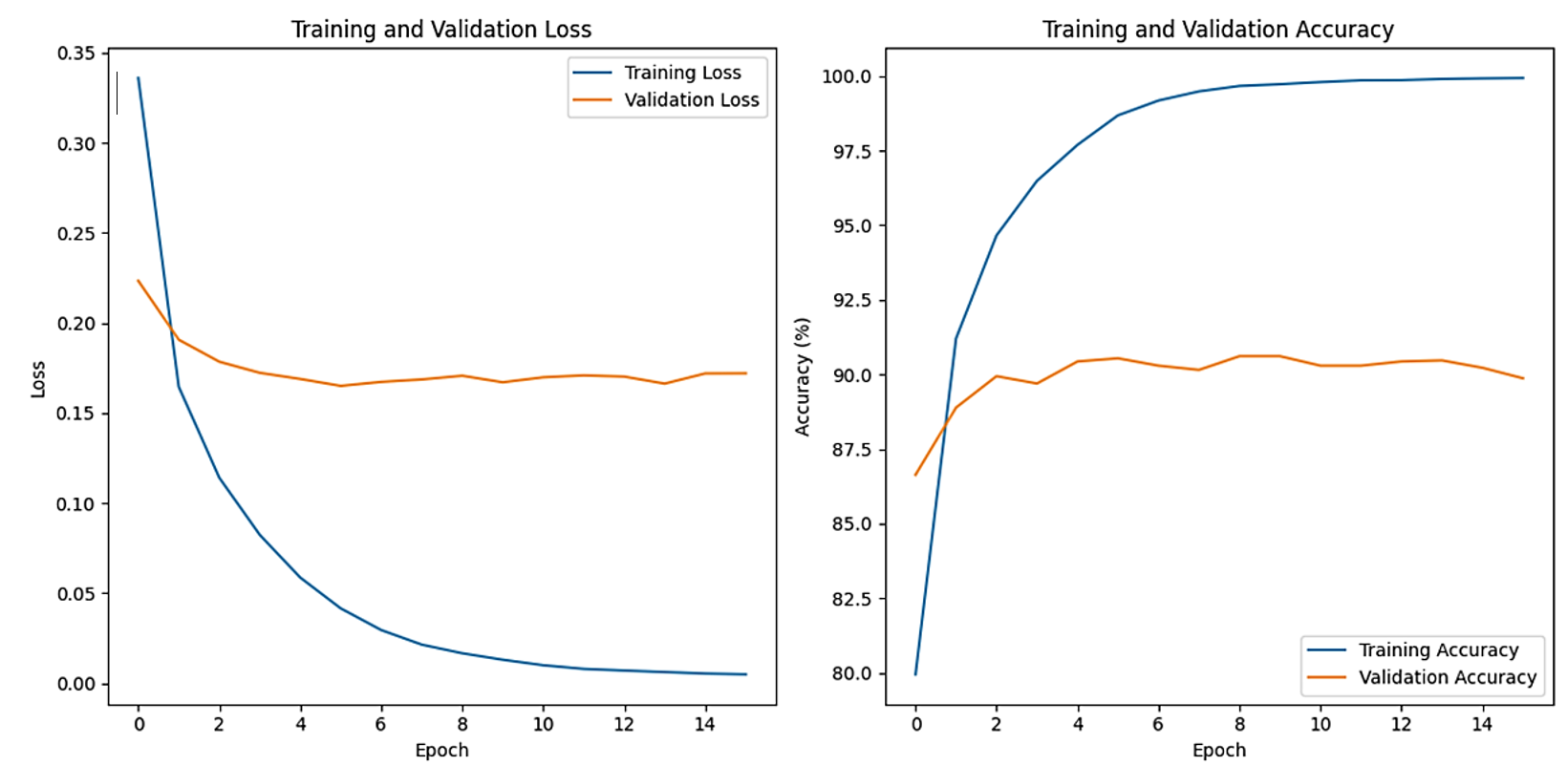}
    \caption{Loss and accuracy curves of the frozen DenseNet121 with the capsule network.}
    \label{fig:frozen_loss_acc}
\end{figure}

\begin{figure}[!ht]
    \centering
    \includegraphics[width=1\linewidth]{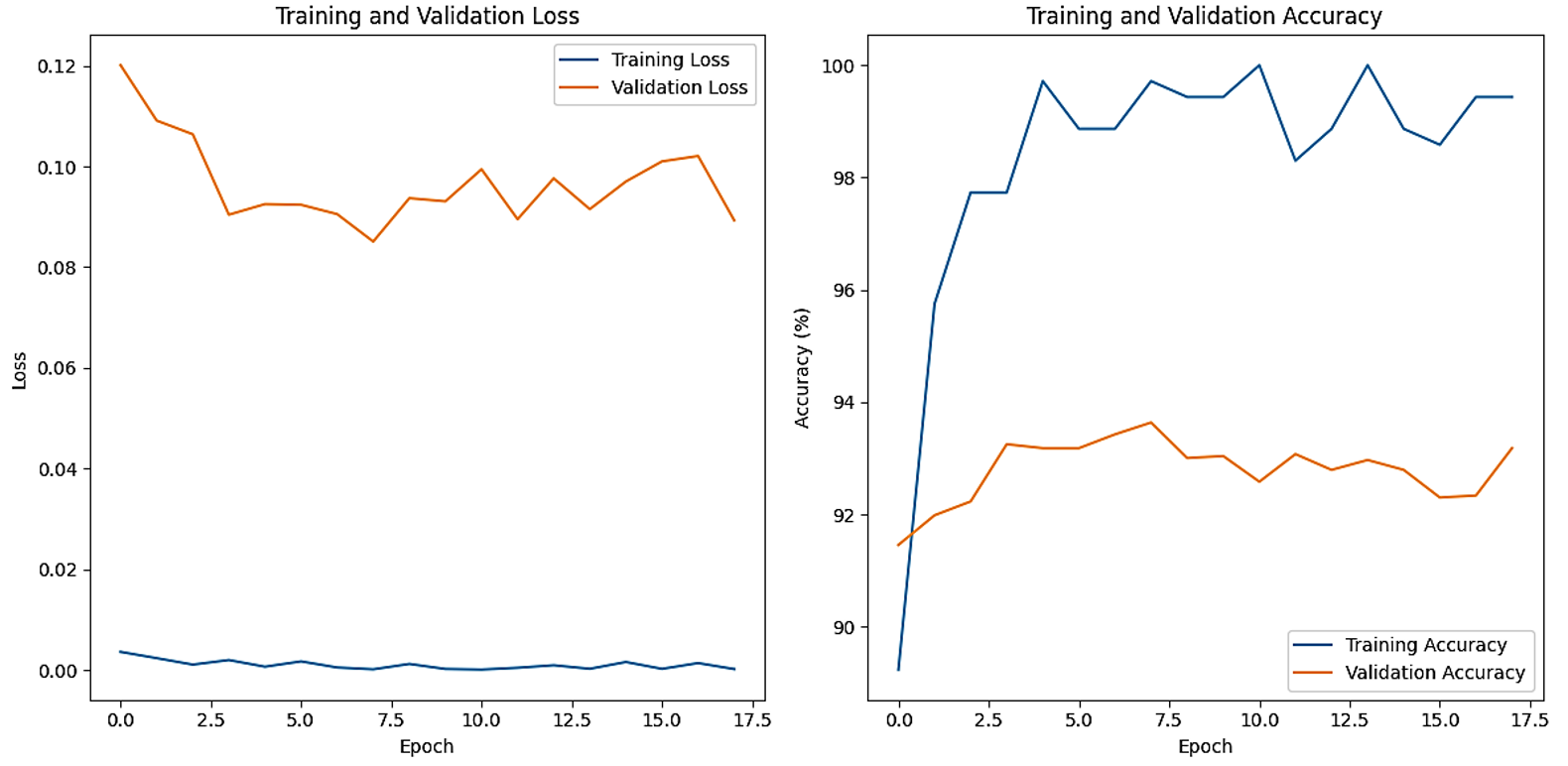}
    \caption{Loss and accuracy curves of the unfrozen DenseNet121 with the capsule network.}
    \label{fig:unfrozen_loss_acc}
\end{figure}

\bibliographystyle{splncs03}
\bibliography{references}

\end{document}